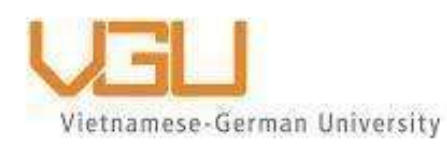

# ENHANCING SHRIMP DISEASE DETECTION VIA DEEP LEARNING AND DATA REFINEMENT FOR RESILIENT AQUACULTURE

Vinh Canh-Thanh Truong
Computer Science Program
Vietnamese-German University
Ho Chi Minh City, Vietnam
15766@student.vgu.edu.vn

Hai-Binh Pham
Computer Science Program
Vietnamese-German University
Ho Chi Minh City, Vietnam
10421006@student.vgu.edu.vn

Ngoc Hong Tran*
Computer Science Program
Vietnamese-German University
Ho Chi Minh City, Vietnam
ngoc.th@vgu.edu.vn
*: *Corresponding Author*

**Abstract:** Shrimp diseases continue to cause devastating losses in the aquaculture industry, driving a critical need for robust, automated detection. This work contributes the first application of Vision Transformers (ViT) and Self-Supervised Learning (SSL) to the shrimp farming domain, addressing both performance bottlenecks and data labeling challenges. We propose two deep learning pipelines to classify four key diseases: Healthy, Black Gill (BG), White Spot Syndrome Virus (WSSV), and a co-infection of both using a dataset of 4,348 images. First, our supervised transfer-learning approach leverages ImageNet-pretrained ViT-Small/16 and EfficientNet backbones. Second, we introduce a contrastive learning framework (SimCLR) with a ViT-Small encoder to extract robust representations from unlabeled images prior to fine-tuning. Our results establish strong new baselines for sustainable aquaculture monitoring. The supervised approach achieves an outstanding 96% accuracy with fast convergence, outperforming traditional generic models, while the label-efficient SSL approach reaches a highly competitive 85% validation accuracy

## 1. Introduction

Shrimp are among the most valuable farmed species within this surge, especially in Asia's intensive systems **[1]**. Disease remains one of the highest risks in the industry **[2]**. In particular, White Spot Syndrome Virus (WSSV) and Black Gills (BG) alone have been associated with recurring shocks to production and revenue. Recent evolution in Computer Vision (CV) has proven that robust performance in ponds is non-trivial **[3]**. Convolutional Neural Network (CNN) **[4]** and Vision Transformer (ViT) **[5]** have been giant players in the field of image classification **[6]**. This highlights the ability to capture long-range dependencies in images, which is crucial to identifying subtle disease symptoms in shrimp **[7]**. Hence, ViT has the potential to enhance the accuracy compared to CNN models.

Furthermore, Self-Supervised Learning (SSL) helped address two common issues of current image classification: dependence on a large volume of data and high labeling costs **[9]**. Suggesting that this technique can be applicable to underwater shrimp images for the ViT encoder to detect the diseases efficiently.

Therefore, in this work, we perform shrimp disease classification following the two proposed approaches:

- **Approach 1:** with Supervised Learning: fine-tunes the pre-trained ImageNet ViT model on a labeled dataset to identify the performance of ViT on a labeled shrimp dataset. Then compared with a popular CNN variance: EfficientNet, also pre-trained on ImageNet **[12]** to highlight the performance between two models on the same labeled dataset.
- **Approach 2:** with Self-supervised Learning (CL to ViT): uses CL with ViT as a backbone encoder to perform training on the unlabeled dataset, we then fine-tune on a labeled dataset with the result model from the previous pretraining process with same backbone ViT model as the encoder.

Our final goal is to measure and compare the result of these two learning approaches to showcase the efficiency of each approach in the classification task with the same dataset.

The remainder of this paper is organized as follows. Section 2 presents related works and analyses their pros and cons. Section 3 introduces the two proposed approaches, that is, supervised learning and self-supervised learning. Section 4 contains the data source which is collected and augmented. The result is demonstrated in Section 5 for both approaches. Finally, we conclude our work and outline future research in Section 6.

## 2. Literature Review

K-means is an early approach for the detection of WSSV lesions. This study by Nagalakshmi in 2013 **[13]** first isolated white lesions on the body before applying the algorithm. Despite its simplicity, it is highly sensitive to environmental conditions such as lighting, camera angle, and pond background, limiting its robustness in real-life scenarios.

With the rise of CNN, a representative study in Vietnam **[14]** trained several standard CNNs on shrimp images contributed by farmers. Despite the non-standard data acquisition scenario, the results yielded a precision around 90% across multiple disease categories. This approach demonstrates that generic CNN backbones could provide a significant improvement over hand-crafted pipelines. More recent CNN variants, such as DenseNet **[15]**, and Inception **[16]**, have further improved the robustness and precision of pond imagery **[17]**. Moreover, the YOLO architecture **[18]**, commonly used for object detection, is also employed. A study by Xu in 2024 **[19]**, using YOLOv8 with multiple features of images extracted, achieving the accuracy of 90%.

However, existing deep learning studies in this domain share two primary limitations. First, CNN-based methods rely on localized convolutional kernels, which can struggle to capture the global, long-range dependencies and distance between pixels within an image **[4]**. Second, these current models are fully-supervised, meaning their performance depends heavily on large, meticulously labelled datasets, resulting in high labeling costs that are difficult to scale in aquaculture.

Our work addresses these gaps. Unlike prior studies restricted to CNN architectures, we introduce Vision Transformers (ViT) **[5]** to the shrimp farming domain. ViTs excel at capturing long-range image dependencies and have successfully achieved nearly 93% accuracy in related fields like fish disease detection **[8]**. Furthermore, to overcome the data dependency bottleneck of supervised approaches, we utilize Contrastive Learning (CL), a Self-Supervised Learning (SSL) technique. While CL has improved performance on limited datasets in medical and plant imaging **[11]**, it remains unexplored for shrimp images. By proposing a ViT backbone integrated with a SimCLR framework, our work provides a label-efficient baseline that explicitly improves current requirements in both accuracy and rigid annotated data of existing studies.

## 3. Methodology

In this section, we illustrate the two proposed approaches that we employed for the current shrimp problem.

### 3.1. Supervised Learning

The objective is to map the input images of the shrimps to categorical labels. Each sample in the dataset is neatly organized in 1 of the 4 classes (Healthy, BG, WSSV, WSSV BG)

**a) EfficientNet**

EfficientNet **[12]** is a popular CNN variance, whose pre-trained model achieves a well-balanced performance on ImageNet **[20]**. In this work, the specific model chosen is EfficientNet B0 with $\phi = 0$ as the backbone encoder. The training is followed by a two-phase approach as follows. Phase 1 is performed with Linear Probing. This phase is optional. In this phase, EfficientNet backbone is frozen and only the classification head is trained for 10 epochs with AdamW Optimizer **[21]**, lr = 1e - 3 and weight decay = 1e - 4. Phase 2 is performed with Fine-tuning. The entire network is unfrozen and optimized. Updating the learning rate = 1e - 4 and weight decay value similar to that of phase 1.

**b) Vision Transformer (ViT)**

In this study, the variant vit_small_path16_224 variant from the timm library with pre-trained weights **[22]** is utilized. The model with attached classification head is then trained for 50 epochs with AdamW as optimizer and Cosine Annealing **[21]** as schedule.

### 3.2. Self-Supervised Learning (SSL)

SimCLR framework **[23]** is a popular recipe to implement CL on a large scale.

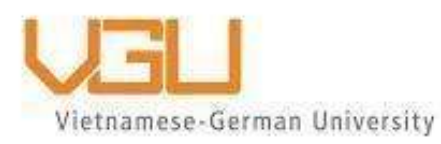

We choose to use the original version of SimCLR among many other versions such as DeepSet SimCLR **[24]** and G-SimCLR **[25]**, due to its simplicity, flexibility, and efficiency. The implementation is composed of two main steps. First, pre-trained with SimCLR framework with ViT-Small as encoder on unlabeled shrimp dataset with 200 epochs. Second, fine-tune the pre-trained result in step 1 on a labeled shrimp dataset with 4 class labels: Healthy, WSSV, BG and WSSV BG with 100 epoch

### 3.3. SimCLR Settings

There are multiple major components utilized in our experiment for SimCLR. They are multiple view image segmentation, a shared encoder, projection head, contrastive (InfoNCE) loss, and normalization and batch. In multiple view image segmentation, for each input, sample two independent augmentations for pre-processing tasks: $x_a$, $x_b$ (see Fig. 1).

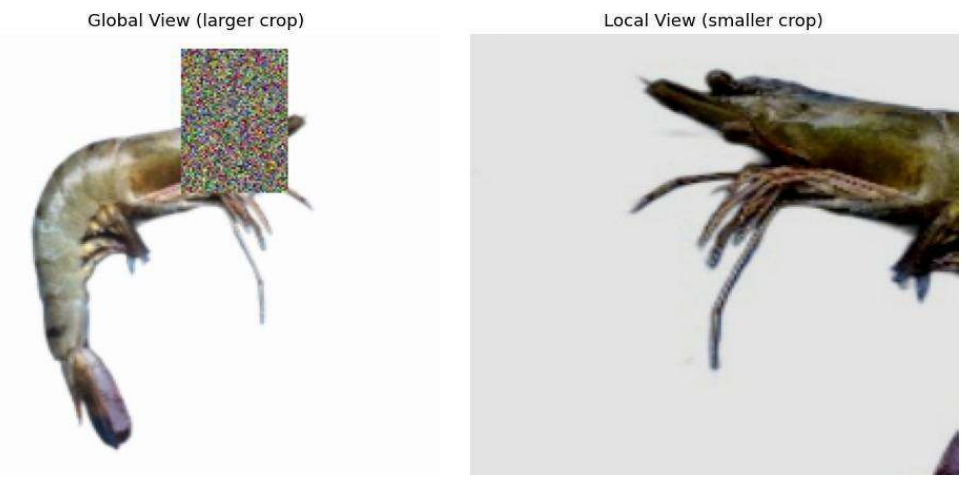

Figure 1. Augmented with Gaussian blur and zoom dual view

In shared encoder, ViT-Small encoder to transform input image to an embedding vector. With projection head, a typical multilayer perceptron (MLP) **[26]**, producing output as a representative embedding vector from input embedding vector after going through the encoder step. Whereas, with contrastive (InfoNCE) loss, for each positive pair maximize their cosine similarity while minimizing similarity to all other samples in the batch. A temperature τ is added to control how peaky the softmax is typically τ = 0.1 as in equation 1 below. Moreover, in normalization and batch, we use L2-normalize before loss.

$$L_i = -log \frac{exp(sim(z_i,z_j)/\tau)}{\sum_{k=1}^{N} 1_{[k\neq i]} exp(sim(z_i,z_k)/\tau} \quad (1)$$

### 3.4. Pre-training SimCLR

Our combining SimCLR and ViT scheme uses a Vision Transformer (ViT) backbone with the standard Patch-Embedding (PE): for each image, two independently augmented views (random resized crop with light color/contrast jitter, blur, small rotations/flips) are created. Each augmented variance is passed through a single 2D convolution Patch Embedding (kernel = stride = 16) that converts the image into non-overlapping tokens, we then prepend a CLS token, add learnable positional embeddings, and pass the sequence through 12 pre-norm Transformer blocks, which represent multi-head self-attention **[27]** and a 2-layer MLP projector (384, 512, 128 respectively). The final LayerNorm CLS vector is the image representation. Training is optimized with InfoNCE loss **[28]** with temperature approximately 0.1 over all 2N views in the batch (see Fig. 2).

### 3.5. Fine-tuning Self-Supervised Learning

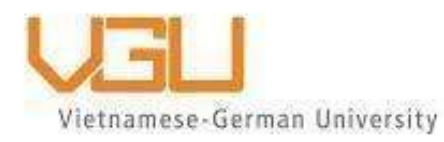

For the 4-class downstream classification (Healthy, BG, WSSV, WSSV BG), we discard the projector (the projection head in the first step) while keeping the encoder of ViT backbone. We also attach a new classification head: Dropout (0.3), Linear (384, 512), ReLU, Dropout (0.2) and Linear (512, 4), respectively. Then, we train with cross-entropy loss in 4 classes.

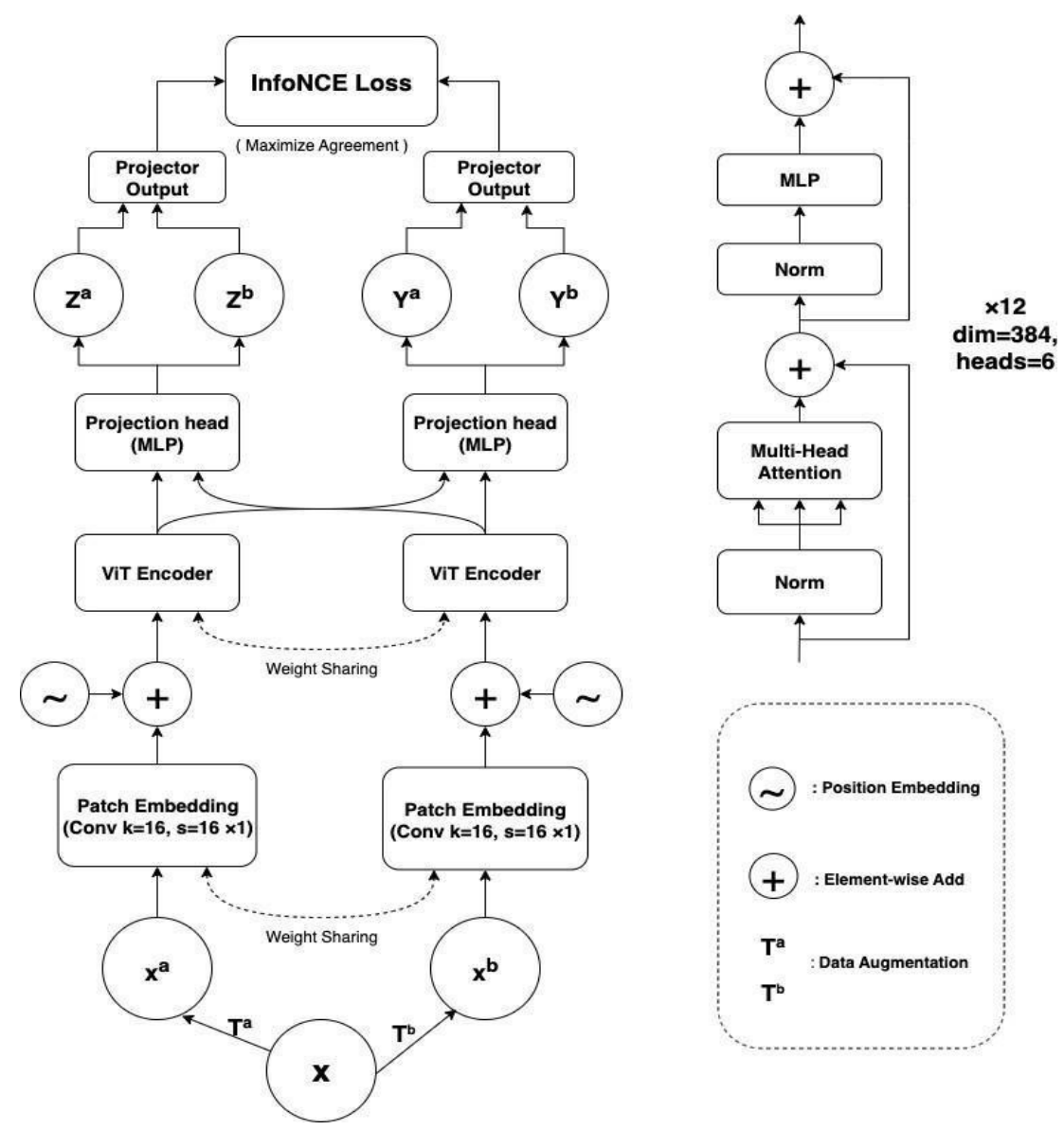

Figure 2. Model architecture with SimCLR using ViT backbone

## 4. Data source

### 4.1. Data description

Our study's data is built and extended primarily on ShrimpDiseaseImageBD **[29]**, a public dataset released by Mendeley Data containing 1,149 raw images organized into four classes: Healthy (403), BG (198), WSSV (328) and (220), plus a companion folder of annotated diseased shrimp with image–label pairs for BG, WSSV, and co-infection. The dataset is published under CC BY 4.0, which allows reuse with attribution. A companion short article in ScienceDirect describes the same four-class structure.

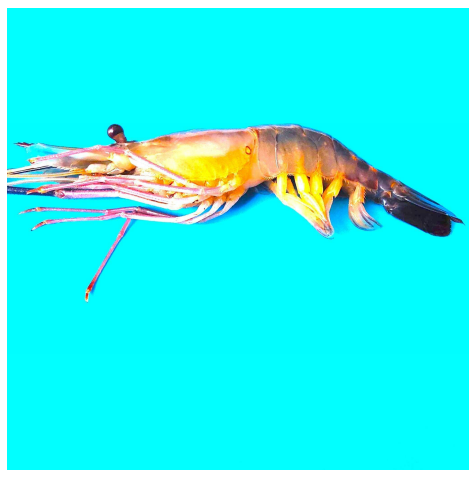

Figure 3. Brightness and contrast shrimp single

### 4.2. Data pre-processing and augmentation

To improve robustness without erasing disease cues, we applied a lightweight augmentation recipe during training. First, we handle the background of images by removing background on a subset of images and mix with originals to reduce spurious background correlations. Then, we apply photometric jitter with the modest adjustments in color, brightness, and contrast (see Fig. 3). After that, we process geometric jitter via random rotation and flips (see Fig. 4). Finally, we do zooming, we perform a new image with zooming the shrimps' issued part.

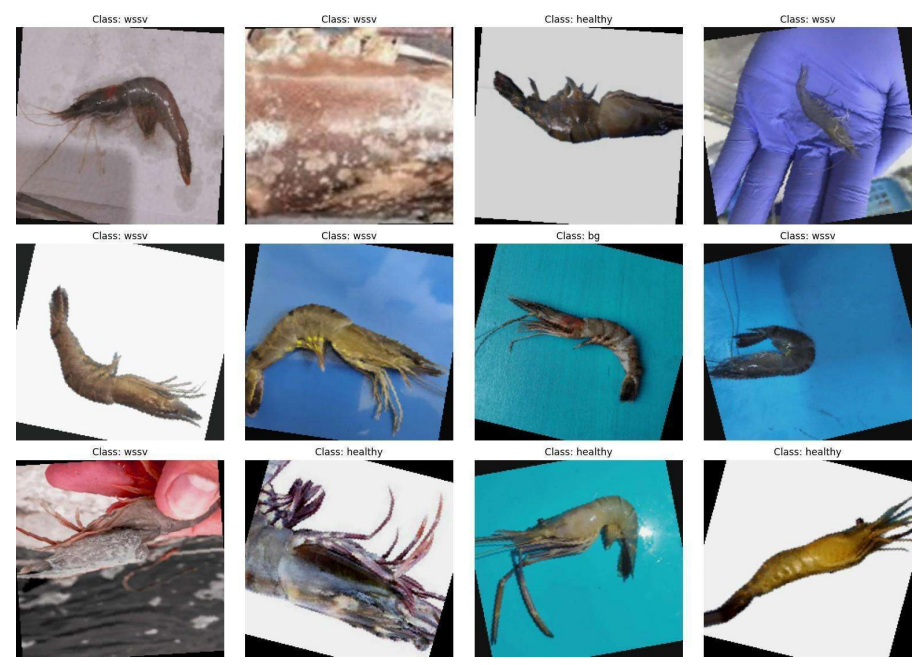

Figure 4. Rotation and flip shrimp in grid view

### 4.3. Final dataset

After curation, augmentation, and quality checks, our working dataset contained 4, 348 images, split train/val/test = 3,421 / 857 / 70 (80/20 split for train/val, with a small holdout test). The classes contained for the train set were Healthy (1,021), BG (558), WSSV (1,266), and WG BG (576); corresponding proportions were preserved as closely as possible in validation and test.

## 5. Experimental Results

### 5.1. Supervised Learning

#### a) EfficientNet

For 10 epochs in phase 1 with the linear probing approach, accuracy improved significantly from roughly 60% from the initial epoch to approximately 86% at the end of the phase. During phase 2 (fine tuning), the accuracy of the training improved to 98.65% with early stopping triggered at epoch 17 and approximately 96% for validation (see Fig. 5).

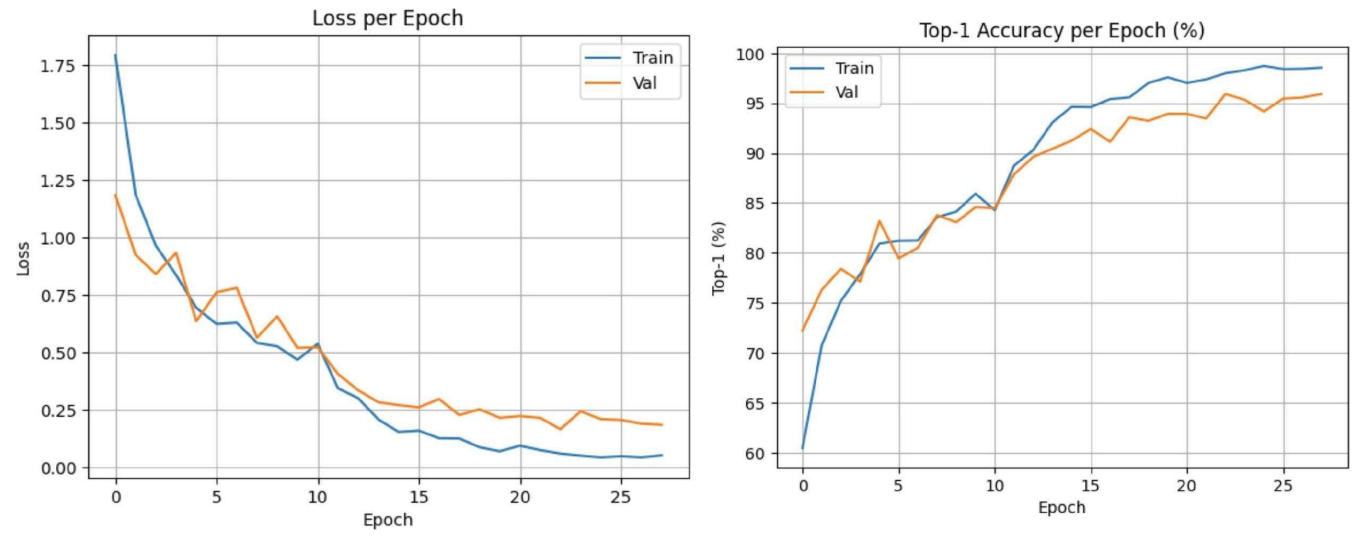

Figure 5. Loss and Accuracy Graph for EfficientNet B0

#### b) ViT

The training process shows a strong and stable learning trajectory. Both training and validation losses decrease rapidly, with the most improvement in the first six epochs. Validation loss minimizes around Epochs 7–8, with slight fluctuation later, suggesting mild overfit. Training accuracy reaches ~99%, while validation accuracy converges closely at 95–96%, indicating robust generalization. The small gap between the curves confirms consistent performance. In summary, ViT-Small achieves efficient convergence and high predictive accuracy, with optimal performance before

overfitting (see Fig. 6)

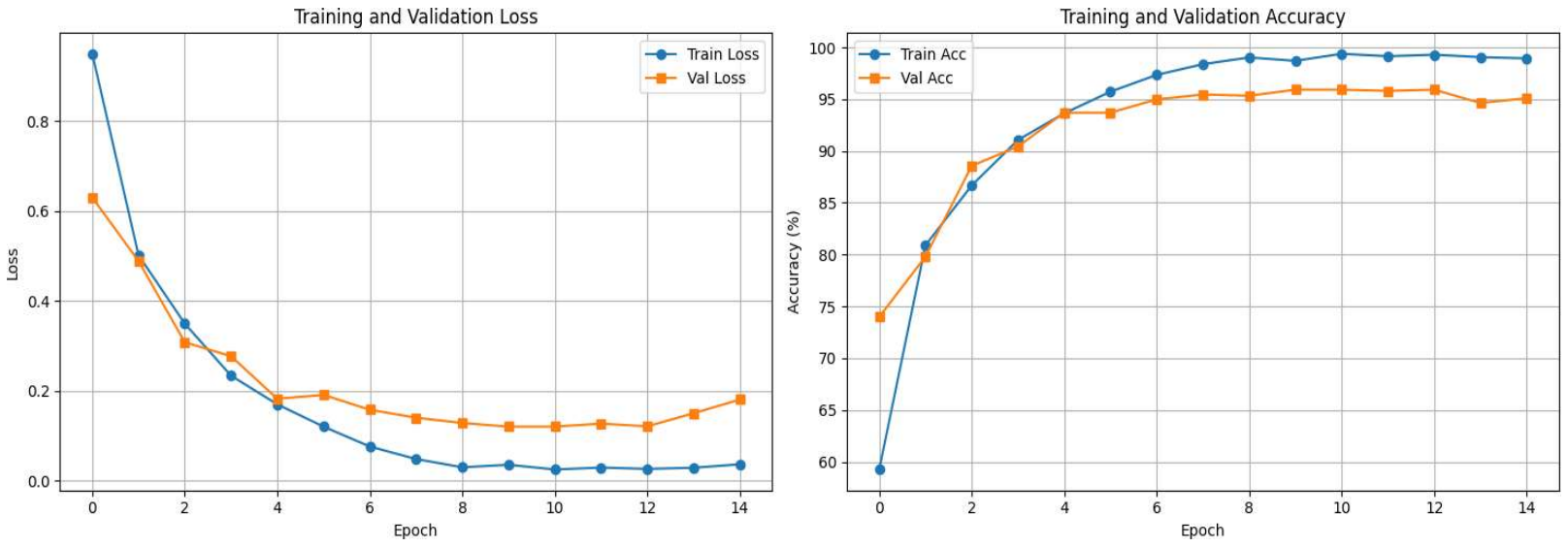

Figure 6: Loss & Accuracy Graph for ViT

### 5.2. Self-supervised Learning SimCLR

#### a) Pre-train

During pre-training, the InfoNCE loss steadily decreased from more than 3.0 to less than 0.3 over 200 epochs, indicating stable convergence of the contrastive framework (see Fig. 7).

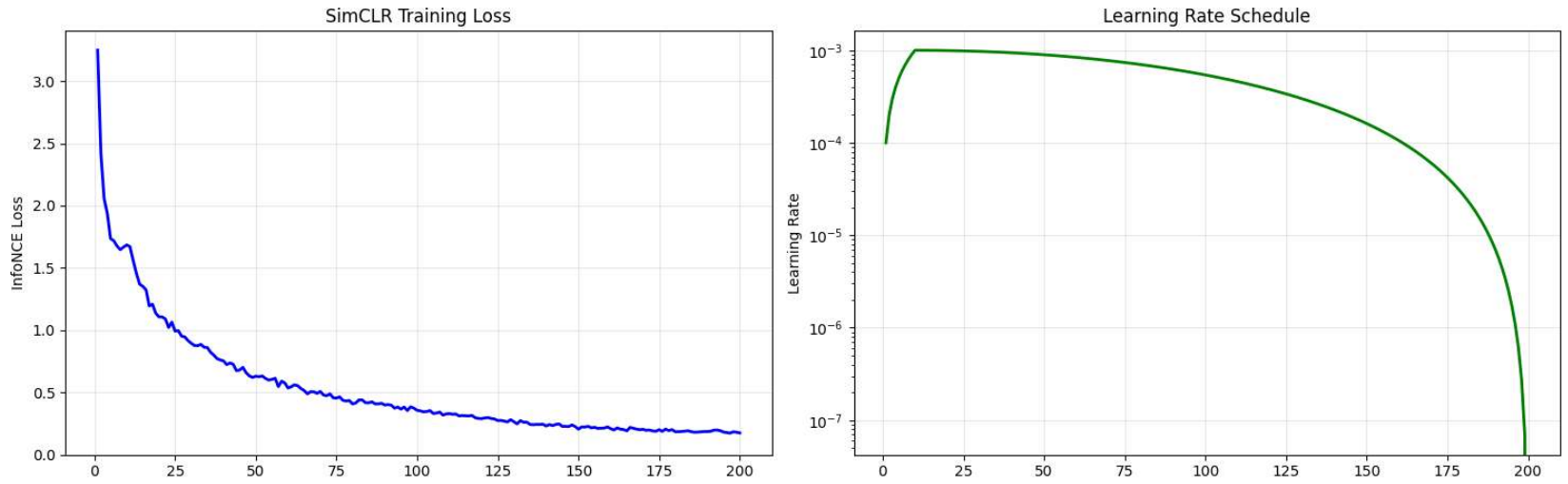

Figure 7: Loss for the pre-training SimCLR with ViT

The t-SNE visualization of the learned representations reveals well-formed and clearly separated feature clusters, demonstrating that the encoder captured semantical structures in the dataset without supervision, which is likely to group similar images based on visual similarities like shrimp orientation, lighting conditions, or subtle disease indicators while pushing dissimilar ones apart (see Fig. 8).

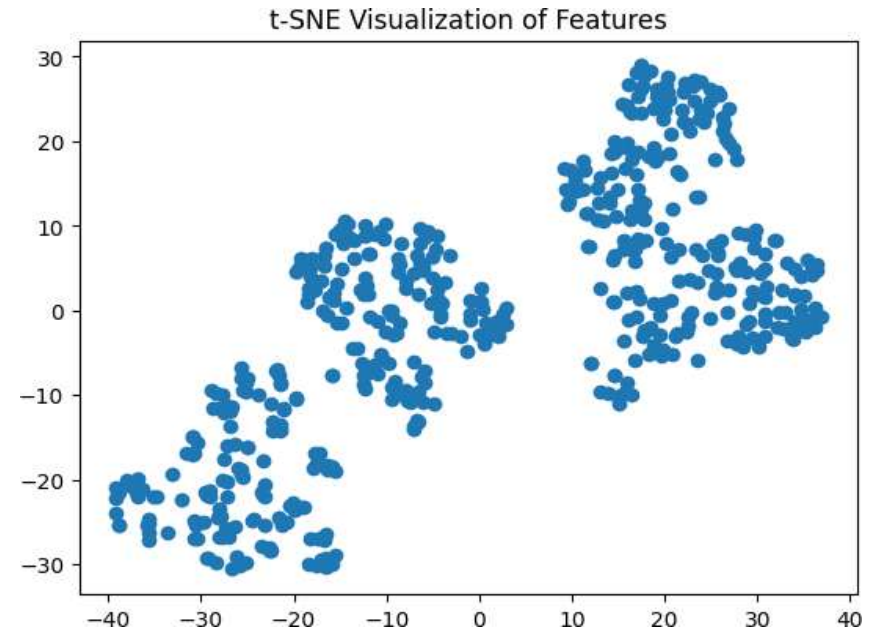

Figure 8: t-SNE visualization of the pre-train SimCLR

#### b) Fine-tuned the pre-trained SimCLR

After pretraining with SimCLR, the Vision Transformer backbone was fine-tuned for 100 epochs. Both training and validation losses decreased steadily, with validation loss stabilizing around 0.5, indicating effective transfer of pre-trained representations. Consequently, the training accuracy was above 85%, while the validation accuracy fluctuates near 80%, suggesting a good generalization performance. The cosine learning rate schedule facilitated smooth convergence by gradually reducing the learning rate during training. The gap between training and validation loss shows a moderate upward trend, little over- fitting at the later epoch, but overall the model retained strong validation performance (see Fig. 9).

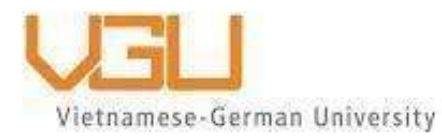

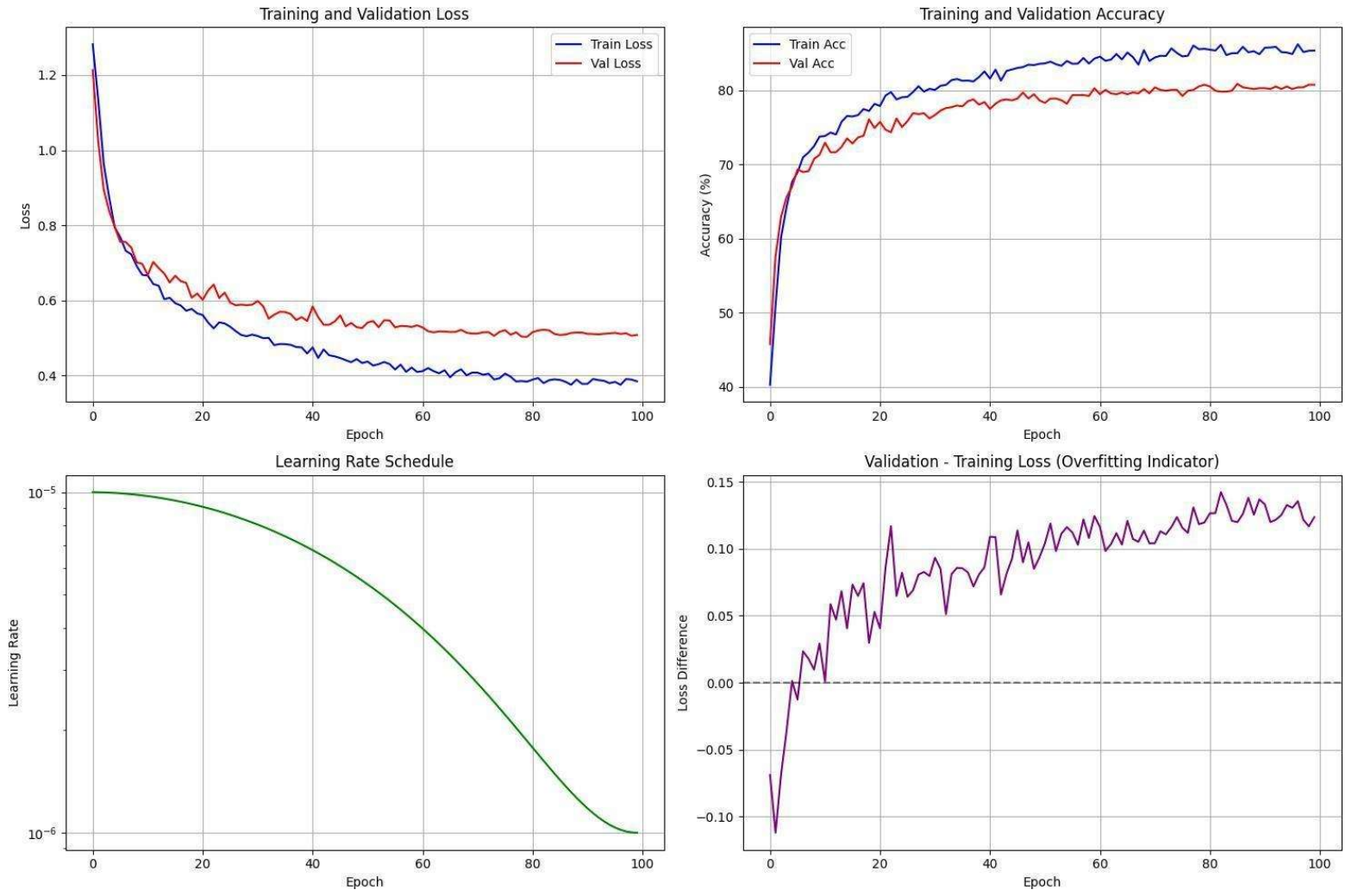

Figure 9: Train and Validation Loss

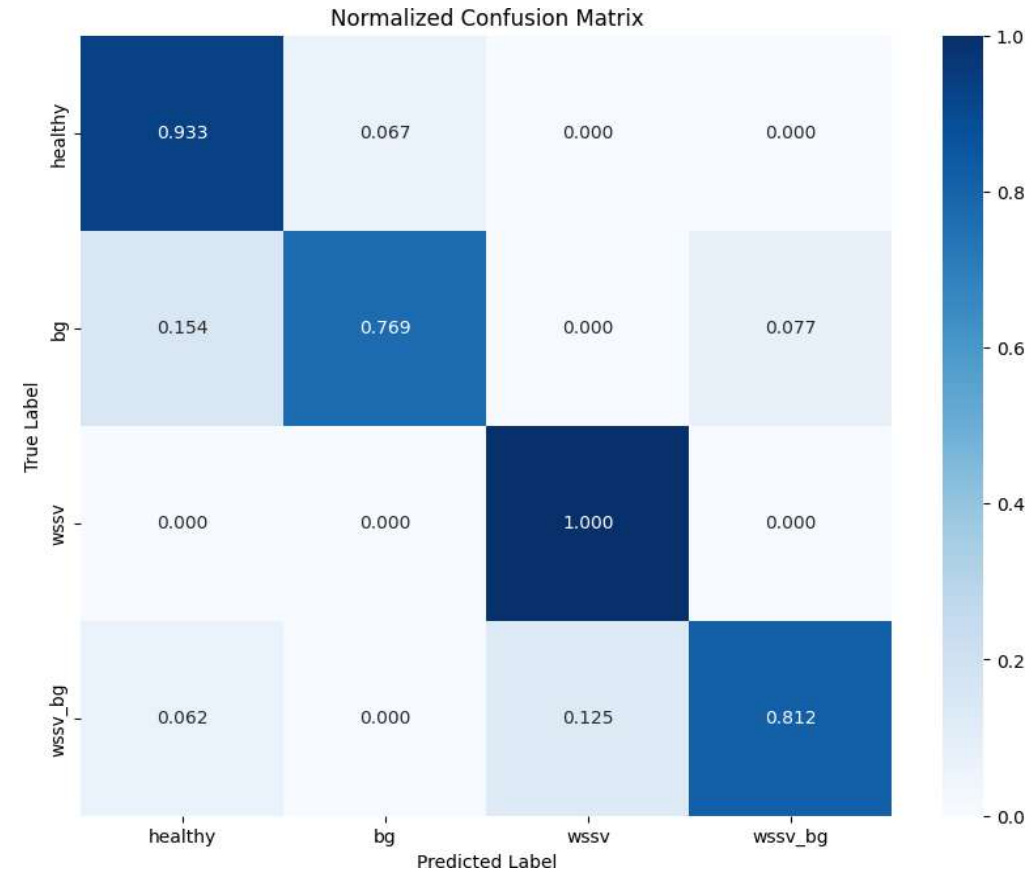

Figure 10: Normalized confusion matrix

The normalized confusion matrix (see Fig. 10) demonstrates the classification performance on a test set across four classes. The model achieved strong discrimination, with recall values of 93.3% for Healthy, 76.9% for BG, 100% for WSSV and 81.2% for WSSV_BG.

**c) Analysis on fine-tuning misclassified results.**

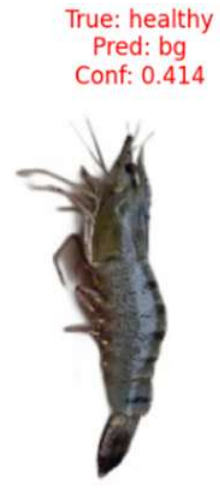

Figure 11: Misclassification Sample Between Healthy and BG

The wrong detection primarily occurred between the Healthy and BG classes (6.7% and 15.4% misclassification, respectively). Biologically, early-stage Black Gill disease often presents as a slight browning that visually overlaps with the natural, darker pigmentation found in some healthy shrimp due to genetic variation or diet **[30]**, and the environmental factors such as mud or organic debris. They can accidentally teach the network to associate darkened gill regions with healthy body parts.

Regarding co-infection confusion on the graph, the model occasionally misclassified WSSV_BG as a single-infection class. This points to a dominance issue where WSSV produces highly salient, high-contrast white spots **[31]** that can overwhelm the network's attention mechanism **[27]**. As a result, the model prioritizes these prominent markers while effectively ignoring the subtler, darker gill features, leading to an incomplete diagnosis.

### 5.3. Discussion

While both approaches demonstrate the viability of ViT architectures, there is a notable performance gap between the fully supervised approach (96% accuracy) and the self-supervised SimCLR approach (85%). This nearly 15% discrepancy can be attributed to the fundamental architecture of contrastive learning and the scale of the dataset.

SimCLR relies on maximizing the agreement between differently augmented views of the same image while minimizing similarity to other samples **[10]**. In this process, the encoder tends to prioritize macro-level features such as shrimp orientation, background color, or overall shape while potentially discarding the fine-grained, micro-level textural anomalies such tiny white spots or subtle gill discoloration that are critical for accurate disease discrimination.

Furthermore, self-supervised frameworks typically require massive quantities of unlabeled data to enrich **[9].** Although our dataset of 4,348 images is big enough for the shrimp domain, it remains relatively small for a contrastive learning paradigm compared to the millions of images used to pre-train the ImageNet baselines on ViT and ImageNet **[12] [22]**.

Consequently, the choice depends on data and resources: when labeled data are sufficient or the computation is tight, the most powerful model is ViT pre-trained on ImageNet, due to the wide range of features and parameters. If unlabeled shrimp images are abundant and labels are scarce, SimCLR and ViT is a compelling alternative to reduce the labeling task.

## 6. Conclusion

This paper represents multiple strategies for the shrimp classification task. By comparing two proposed approaches, that is, supervised learning through the pre-trained models on ImageNet dataset and self-supervised learning with fine-tuning the downstream dataset. Furthermore, future work will expand the test set across farms and imaging conditions with more than 1000 data for each disease pattern. Exploring calibration and thresholding for decision, support and extension of the approach to additional shrimp pathologies, and thus moving toward a robust, field-deployable diagnostic pipeline.